\documentclass[10pt,twocolumn,letterpaper]{article}
\usepackage{bm}

\usepackage{iccv}
\usepackage{times}
\usepackage{epsfig}
\usepackage{graphicx}
\usepackage{amsmath}
\usepackage{amssymb}
\usepackage[algoruled,boxed,lined]{algorithm2e}
\usepackage{mathtools}
\graphicspath{{figures/}}

\usepackage[pagebackref=true,breaklinks=true,letterpaper=true,colorlinks,bookmarks=false]{hyperref}

\usepackage[breaklinks=true,bookmarks=false]{hyperref}

\iccvfinalcopy % *** Uncomment this line for the final submission

\def\iccvPaperID{2266} % *** Enter the ICCV Paper ID here

\ificcvfinal\fi

\begin{document}

%%%%%%%%% TITLE
\title{MIC: Mining Interclass Characteristics for Improved Metric Learning}
% \title{MIC: Mining Interclass Characteristics for Improved Visual Metric Learning}

% \title{HIP: Improving Metric Learning by Harvesting Intraclass Properties}
% \title{MIC: Improving Metric Learning by Mining Intraclass Characteristics}
% \title{MIST: Metric Learning by Mining Intraclass Structure}
% \title{HIS: Metric Learning by Harvesting Intraclass Structure}
% \title{iMEC: Improving Metric Learning by Extracting Intraclass Characteristics}
% \title{EIS: Improving Metric Learning by Extracting Intraclass Structure}

% \author{
% Karsten Roth\thanks{Indicates equal contribution}\and
% Biagio Brattoli$^\ast$\\\\
% HCI/IWR, Heidelberg University, Germany\\
% {\tt\small firstname.lastname@iwr.uni-heidelberg.de}\\
% \and
% Bj\"orn Ommer
% }

\author{Karsten Roth\thanks{Indicates equal contribution}, \space Biagio Brattoli$^\star$, \space Bj\"orn Ommer\\
HCI/IWR, Heidelberg University, Germany\\
{\tt\small firstname.lastname@iwr.uni-heidelberg.de}}

% \and
% Biagio Brattoli$^\star$\\
% {\tt\small biagio.brattoli@iwr.uni-heidelberg.de}
% \and
% Bj\"orn Ommer\\
% {\tt\small biagio.brattoli@iwr.uni-heidelberg.de}
% }

\maketitle
\thispagestyle{empty}

%%%%%%%%% ABSTRACT
\begin{abstract}
%%%%%%%%%%%%%%%%%%%%%%%%%%%%%%%%%%%%%%%%%%%%%%%%%%%%%%%%%%%%%%%%%%%%%%%%%%%%%%%
Metric learning seeks to embed images of objects 
%in such a way that user-defined relations, such as same/different object class, 
% such that user-defined relations like same/different object classes are captured by the embedding space.
such that class-defined relations are captured by the embedding space.
 However, variability in images is not just due to different depicted object classes, but also depends on other latent characteristics such as viewpoint or illumination. In addition to these structured properties, random noise further obstructs the visual relations of interest. 
The common approach to metric learning is to enforce a representation that is invariant under all 
%factor, but the one of interest, 
factors but the ones of interest.
%so they can be neglected. 
In contrast, we propose to explicitly learn the latent characteristics that are shared by and go across object classes. We can then directly explain away 
%the structured visual variability, 
structured visual variability,
rather than assuming it to be unknown random noise.

We propose a novel surrogate task to learn visual characteristics shared across classes with a separate encoder. This encoder is trained jointly with the encoder for class information by reducing their mutual information. On five standard image retrieval benchmarks the approach significantly improves upon the state-of-the-art. Code is available at \href{https://github.com/Confusezius/metric-learning-mining-interclass-characteristics}{https://github.com/Confusezius/metric-learning-mining-interclass-characteristics}.
\end{abstract}

%%%%%%%%% BODY TEXT
% \blue{Leveraging these characteristics can help Computer Vision applications to learn features that solve a given task. Looking at object classification, these tasks revolve around the model finding discriminative characteristics (s.a. car shapes) to group images according to predefined classes. 
% However, to resolve this perfectly, the model needs to understand how objects within a class can behave without leaving said class. Untargeted, this problem relates to a high intra-class variance that models target by treating the variability as false noise in the training data. These issues of high intra-class variance affect more complex tasks such as metric learning especially heavy.} 
%Computer Vision models leverage these image structures to learn the specific

\section{Introduction}
Images live in a high dimensional space rich of structured information and unstructured noise. Therefore an image can be described by a finite combination of latent characteristics. The goal of computer vision is then to learn the relevant latent characteristics needed to solve a given task. %, while simultaneously learning to be invariant to noise. 
% Particularly in object classification, the model extracts the discriminative characteristics (e.g. car shape) to group the images according to predefined classes. To do so successfully, the large intra-class variability needs to be tackled. Modern classifiers can easily learn to be invariant to unstructured noise (e.g. random clutter, occlusion, image brightness).
Particularly in object classification, discriminative characteristics (e.g. car shape) are used to group the images according to predefined classes. To tackle the intra-class variability, modern classifiers can easily learn to be invariant to unstructured noise (e.g. random clutter, occlusion, image brightness). However, a considerable part of the variability is due to structured information shared among classes (e.g. view points and notions of color)%, which, following this procedure, will be disregarded as noise.
% Thus, instead of getting rid of the structured information, a better alternative is to explain them away, leaving only true noise as variability within classes and therefore producing a model more robust to intra-class variance.

\begin{figure}[t]
\begin{center}
   \includegraphics[width=0.99\linewidth]{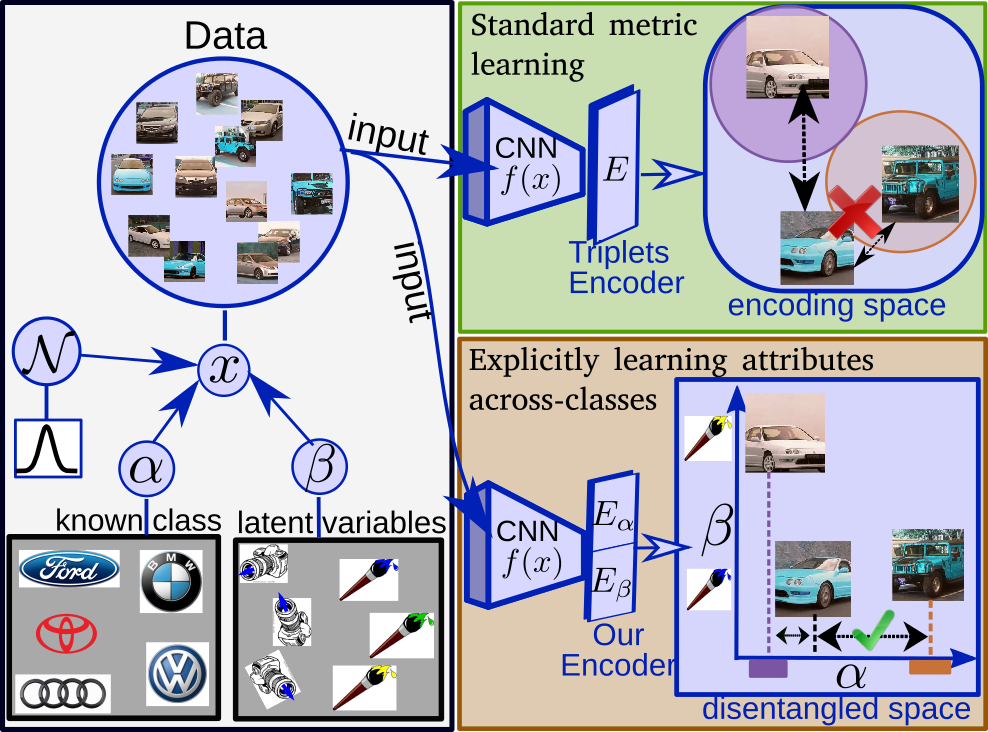}
\end{center}
   \caption{(Left) Images can be described by combinations of latent characteristics and white noise. (Green) Standard metric learning encoders extract class-discriminative information $\alpha$ while disregarding object-specific properties $\beta$ (e.g. color, orientation). Achieving invariance to such characteristics requires substantial training data. (Brown) Instead, the model can explain them away by learning their structure explicitly. Our novel approach explicitly separates class-specific and shared properties during training to boost the performance of the discriminative encoding.}%In the example, the blue car is wrongly classified because only white cars are available from the same class...}%
\label{fig:page1}
\end{figure}

\begin{figure*}[t]
\begin{center}
   \includegraphics[width=0.90\linewidth]{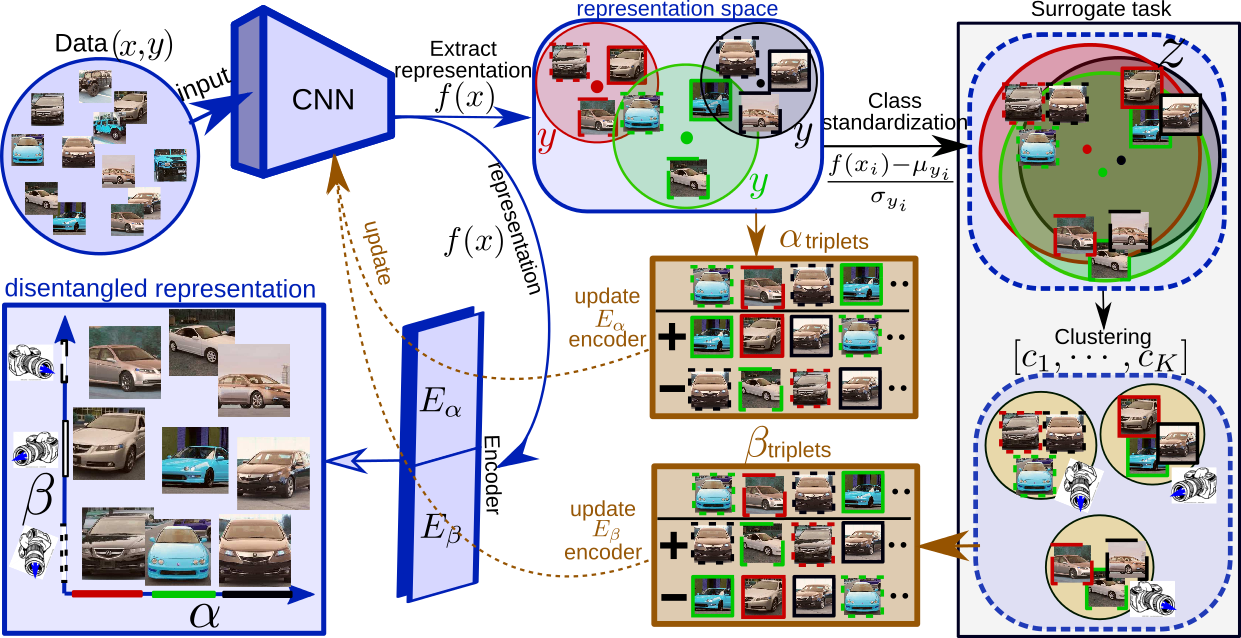}
\end{center}
%   \caption{\textbf{Overview of our approach}. %A metric learning encoder benefits from explaining away the structure information shared among classes.
%   We aim to learn two separate encoding spaces such that the class information $\alpha$ extracted by $E_\alpha$ is free from shared properties $\beta$ by explicitly describing them through an auxiliary encoder $E_\beta$.
%   Given a set of image/label pairs $(x,y)$, a CNN extracts the feature representation $f(x)$, in which images are grouped by class specific (car model) and shared (orientation, color) characteristics. The ground-truth labels (\textit{boundary color}) are used to train the class encoder $E_\alpha$ using a triplet-based loss function. Simultaneously, our auxiliary encoder $E_\beta$ is trained using labels taken from a surrogate task (right). To provide surrogate labels free of class-discriminative information, we standardize the representation per class. As clustering mines the structure $\beta$ remaining in the data representation (contour line-styles), $E_\alpha$ learns a robust, $\beta$-free encoding, which is now explicitly explained by $E_\beta$.
% }
   \caption{\textbf{Overview of our approach}. %A metric learning encoder benefits from explaining away the structure information shared among classes.
   We aim to learn two separate encoding spaces s.t. class information $\alpha$ extracted by $E_\alpha$ is free from shared properties $\beta$ by explicitly describing them through an auxiliary encoder $E_\beta$.
   Given a set of image/label pairs $(x,y)$, their CNN feature representation $f(x)$ groups images by both class specific (car model) and shared (orientation, color) characteristics. %
   We separate these by training the class-discriminative encoder $E_\alpha$ with ground-truth labels (\textit{boundary color}). Simultaneously, an auxiliary encoder $E_\beta$ is trained on labels from a surrogate task (right) to explain away interclass features. The required surrogate labels are generated by standardizing the embedded training data per class and performing clustering. This recovers labels representing the shared structures $\beta$ (\textit{contour line-styles}). Training both tasks together, $E_\alpha$ learns a robust, $\beta$-free encoding, which is now explicitly explained by $E_\beta$.
   }
\label{fig:model}
\end{figure*}

% The high intra-class variance affects more complex tasks, like metric learning.
% More complex tasks such as metric learning are affected by high intra-class variance as well.
% Metric learning is very useful since the lack of predefined labels enables the comparison of classes never seen during training.

% Unlike classification where images are associated with categorical labels, metric learning approaches learn to define similarities between images based on their properties, especially useful when comparing classes never seen during training.
For metric learning this becomes especially important. As metric learning approaches project images into a high-dimensional feature space to measure similarities between images, every learned feature contributes. This means that finding a strong set latent characteristics is crucial.
% However, for a detailed comparison between images, a highly descriptive set of latent characteristics needs to be learned. Unfortunately, this leads to metric learning being particularly vulnerable towards intra-class variability\cite{dvml}.
Learning the characteristics shared across classes should therefore benefit the model \cite{dvml}, as it can better explain the object variance within a class. 
Take for example a model trained only on white cars of a certain category. This model will very likely not be able to recognize a blue car of the same category (Fig.\ref{fig:page1} top-right). 
In this example, the encoder ignores the concept of "color" for that particular class, even though it can be learned from the data as a latent variable shared across all cars (Fig.\ref{fig:page1} bottom-right). 
This is a typical generalization problem and is traditionally solved by providing more labeled data. However, besides being a costly solution, metric learning models need to also generalize to unknown classes, a task which should work independently from the amount of labels provided.

Explicitly modeling intra-class variation has already proven successful\cite{dvml,spatialtransform,gstrs}, %Pooling layers for example directly model the possible translation of an object in the image, while 
such as spatial transformer layers \cite{spatialtransform}, which explicitly learn the possible rotations and translations of an object category. 

We therefore propose a model to discriminate between classes while simultaneously learning the shared properties of the objects. 
To strip intra-class characteristics away from our primary class encoder, thereby facilitating the task of learning good discriminative features, we utilize an auxiliary encoder.
While the class encoder can be trained using ground-truth labels, the auxiliary encoder is learned through a novel surrogate task which extracts class-independent information without any additional annotations.
% Since shared and discriminative characteristics are orthogonal to each other, a single embedding hardly solves the problem.
% While the class encoder can be trained using ground-truth labels, the auxiliary encoder learns by a novel surrogate task which extracts class-independent information without any additional annotations.
%Thus, we introduce an automatic metric learning approach to separate these properties into two encoding spaces, without using any additional annotations(Fig.\ref{fig:page1}). This allows us to solve both tasks simultaneously.
Finally, an additional mutual information loss further purifies the class encoder from non-discriminative characteristics by eliminating the information learned from the auxiliary encoder.%mutual information between the two encodings to be minimal.

This solution can be utilized with any standard metric learning loss, as shown in the result section. Our approach is evaluated on three standard benchmarks for zero-shot learning, CUB200-2011 \cite{cub200-2011}, CARS196 \cite{cars196} and Stanford Online Products \cite{lifted}, as well as two more recent datasets, In-Shop Clothes \cite{inshop} and PKU VehicleID \cite{pku}. The results show that the proposed approach consistently enhances the performances of existing methods.

\begin{algorithm}[t]
\label{alg:algorithm}
\caption{Training a model via MIC}

\SetKwFunction{Cluster}{Cluster}
\SetKwFunction{Embed}{Embed}
\SetKwFunction{Forward}{Forward}
\SetKwFunction{Split}{Split}
\SetKwFunction{Backward}{Backward}
\SetKwFunction{GetBatch}{GetBatch}
\SetKwFunction{Stand}{Stand}
\SetKwFunction{InterLoss}{ClassLoss}
\SetKwFunction{IntraLoss}{SharedLoss}
\SetKwFunction{Loss}{Loss}
\SetKwFunction{Disent}{Disent}
\SetKwFunction{Finetune}{Finetune}
\SetKwFunction{MSE}{MSE}
\SetKwInOut{Input}{input}
\SetKwInOut{Init}{initialization}
\SetKwInOut{Constants}{parameters}
\SetKwInOut{Output}{output}
\SetKwRepeat{Repeat}{repeat}{until}

\SetAlgoLined
% \Input{
%     Data $X$, Encoder $E$, CNN $f$,
%     Class Targets $Y_\alpha$, Minibatchsize $bs$, Number of Cluster $C$, 
%     Epochs before $Y_\beta$ update $T_U$, disentanglement $\gamma$
%     }\
\textbf{Input:} data $X$, full encoder $E$, inter-/intra class encoders $\{E_\alpha, E_\beta\}$, CNN $f$, class targets $Y_\alpha$,  batchsize $bs$, clusternumber $C$, update frequency $T_U$, (adversarial) mutual information loss $l_d$ and weight $\gamma$, projection network $R$, gradient reversal op $r$, metric learning loss functions for $E_{\alpha,\beta}$ $l_{\alpha,\beta}$
\newline

$Y_\beta \leftarrow$ \Cluster{\Stand{\Embed{$X$, $E$, $f$}}, $C$}

$epoch$ $\leftarrow$ 0

\While{Not Converged}{
    \Repeat{end of epoch}{
    
        $b_\alpha, b_\beta$ $\leftarrow$ \GetBatch{$X$, $Y_\alpha$, $Y_\beta$, $bs$}
        
        ${e}_{\alpha,\beta} \leftarrow$ \Embed{$b_{\alpha,\beta}$, $E_{\alpha,\beta}$, $f$}
        
        %$\mathcal{L}_\alpha \leftarrow$ \Loss{${emb}_\alpha$, $Y_\alpha$}$-$ $\gamma$\MSE{${emb}_\alpha$, ${emb}_\beta$}
        
        $L^\alpha \leftarrow$ $l_\alpha$(${e}_\alpha$, $Y_\alpha$) $+$ $\gamma\cdot l_d$(${e}^r_\alpha$, R(${e}^r_\beta$))
        
        $E_\alpha, f \leftarrow$ \Backward{$L^\alpha$}

        ${e}_{\alpha,\beta} \leftarrow$ \Embed{$b_{\alpha,\beta}$, $E_{\alpha,\beta}$, $f$}
        
        $L^\beta \leftarrow$ $l_\beta$(${e}_\beta$, $Y_\beta$)$+$ $\gamma\cdot l_d$(${e}^r_\alpha$, R(${e}^r_\beta$))
        
        %$\mathcal{L}_\beta \leftarrow$ $\mathcal{L}_\beta$ $-$ $\gamma$\MSE{${e}_\alpha$, ${e}_\beta$}
        
        $E_\beta, f \leftarrow$ \Backward{$L^\beta$}
        
    }
    \If{epoch $mod$ $T_U == 0$}{
        $Y_\beta \leftarrow$ \Cluster{\Embed{X,$E_\beta$,$f$}, $C$}
    }
    $epoch \leftarrow epoch+1$ 
}
%$(Optional)$ $f$, $\theta \leftarrow$ \Finetune{$X$, $T_{inter}$, $f$, $\theta$} 
\end{algorithm}

%-------------------------------------------------------------------------
\section{Related Work}
After the success of deep learning in object classification, many researchers have been investigating neural networks for metric learning. A network for classification extracts only the necessary features for discrimination between classes. 
Instead, metric learning encodes the images into an euclidean space where semantically similar ones are grouped much closer together. %a meaningful distance can be computed between two samples. 
This makes metric learning effective in various computer vision applications, such as object retrieval \cite{lifted,margin}, zero-shot learning \cite{margin} and face verification \cite{chopra2005learning,semihard}. The triplet paradigm \cite{semihard} is the standard in the field and much work has been done to improve upon the original approach. As an exponential number of possible triplets makes the computation infeasible, many papers propose solutions for mining triplets more efficiently \cite{margin,semihard,smartmining,htl,miningmanifold}. Recently, Duan \etal. \cite{daml} have proposed a generative model to directly produce hard negatives. ProxyNCA \cite{proxynca} generates a set of class proxies and optimizes the distance of the anchor to said proxies, solving the triplet complexity problem. 
Others have explored orthogonal directions by extending the triplet paradigm, e.g. making use of every sample in the (specifically constructed) batch at once \cite{lifted,npairs}, enforcing an angular triplet constraint \cite{angular}, minimizing a cluster quality surrogate \cite{facilitylocation} or optimizing the overlap between positive and negative similarity histograms \cite{histogram}.  
In addition, ensembles have been quite successfully used by combining multiple encoding spaces \cite{bier,abier,hdc,freund1997decision} to maximize their efficiency.
% An interesting work from Liu \etal.\cite{dvml} explicitly decompose each image into class-specific and intra-class embedding using a generative model. In a way, our work follows this direction since we also train an encoding to explicitly describe properties which are not discriminative across classes. However, we search for structure shared between classes instead of modeling the intra-class variance per sample and, more importantly, leverage this second encoder to purify the main encoder from indiscriminative features through a mutual information loss.
% \blue{Similarly to \cite{abier}, we also minimize the mutual information between encodings }

Our work makes use of class-agnostic grouping of our data (see e.g. \cite{partialordering,cliquecnn}) and shares similarities with proposals from Liu \etal. \cite{dvml}, who explicitly decompose images into class-specific and intra-class embeddings using a generative model, as well as Bai \etal. \cite{gstrs}, who, before training, divide each image class into subgroups to find an approximator for intra-class variances that can be included into the loss. 
However, unlike \cite{gstrs} and \cite{dvml}, we explicitly search for structures shared between classes instead of modelling the intra-class variance per sample \cite{dvml} or class \cite{gstrs}. In addition, unlike \cite{gstrs}, we assume class-independent intra-class variance and iteratively train a second encoder to model intra-class features, thereby purifying the main encoder from non-discriminative features and achieving significantly better results.
% \blue{Our mutual information loss shares similarities to a proposal by \cite{abier} who in a multi-embedding setup use a small neural network to project one encoding into the space of another, allowing for a meaningful similarity measure independent of encoding dimensions.}
% In a way, our work follows this direction since we also train an encoding to explicitly describe properties which are not discriminative across classes. However, we search for structure shared between classes instead of modeling the intra-class variance per sample and, more importantly, leverage this second encoder to purify the main encoder from indiscriminative features through a mutual information loss.

Finally, some works have exploited the latent structure of the data as a supervisory signal \cite{jigsaw,jigsaw++,deepcluster,ourcvpr,oureccv,ourgcpr,dcesml}. In particular, Caron \etal. \cite{deepcluster} learn an unsupervised image representation by clustering the data, starting from a Sobel filter prior initialization. Our approach includes such latent data structures in a similar way, however we use it as auxiliary information to improve upon the metric learning task.
% \blue{Our training procedure for the second encoder shares similarities to the unsupervised learning setup proposed by Caron \etal.\cite{deepcluster}, who use iterations of training and standard clustering on embedded data to train an ImageNet classifier. While they apply their method directly on mostly unaltered ImageNet data, we utilize specifically transformed data to extract semantically different features.}
% Our work follows this direction, however our model learns to represent the variance not class-dependent but rather for the object category as a whole.

%-------------------------------------------------------------------------
\section{Improving Metric Learning}
The main idea behind our method is the inclusion of class-shared characteristics into the metric learning process to help the model explain them away. In doing so, we would gain robustness to intrinsic, non-disciminative properties of the data, which is contrary to the common approach of simply forcing invariance towards them.
However, three main problems arise with this approach, namely:
% \textit{(i)} Since the class and class-independent characteristics are \blue{orthogonal} to each other, solving the task with a single encoder is inefficient. 
\textit{(i)} Extracting both class and class-independent characteristics using a single encoder is infeasible and detrimental to the main goal.  
\textit{(ii)} We lack the labels for extracting these latent properties.
\textit{(iii)} We need to explicitly remove unwanted properties from the class embedding. We propose solutions to each of these problems in sections \ref{sec:orthenc}, \ref{sec:surrogate} and \ref{sec:loss}.

\begin{figure}[t]
\begin{center}
   \includegraphics[width=0.99\linewidth]{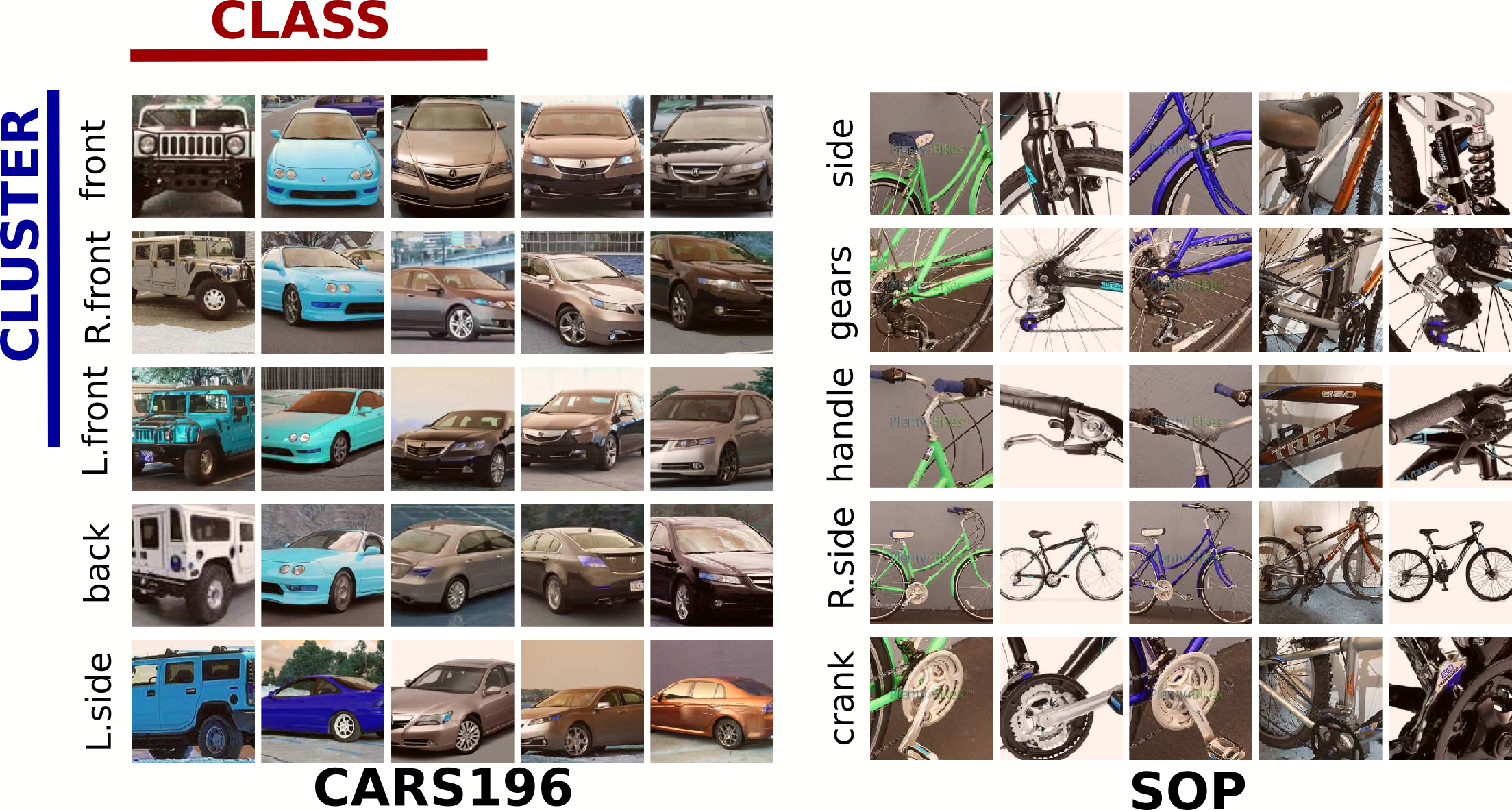}
\end{center}
   \caption{Example of clustering the data based on $Z$ (see Sec\ref{sec:surrogate}) for two datasets: CARS196\cite{cars196} and SOP\cite{lifted}. We group the dataset into 5 clusters (rows) and select the first 5 classes (columns) with at least one sample per cluster. For each entry, we selected the sample closest to the centroid per class. On the left is our interpretation of the cluster structure. The results show that subtraction of the class-specific features by standardization helps to group images based on more generic properties, like car orientation and bike parts.}
\label{fig:clusters}
\end{figure}

% \subsection{Metric Learning}
\subsection{Preliminaries}
% \red{Metric learning encodes the characteristics that better discriminate between classes into an embedding vector. EITHER EXPLAIN ML OR GO TO THE MAIN PROBLEM, THIS IS TOO MUCH IN BETWEEN Meanwhile, shared properties between classes are discarded. When comparing images, the extracted characteristics contribute equally to produce a single similarity score, thus it is particularly crucial for metric learning to get rid of any unwanted information
% The goal of metric learning is to train an encoder $E$ such that the images $x_i$ from the same class $y$ are nearby in the encoding space and samples from different classes are far apart.} 
Metric learning encodes the characteristics that discriminate between classes into an embedding vector, with the goal of training an encoder $E$ such that images $x_i$ from the same class $y$ are nearby in the encoding space and samples from different classes are far apart, given a standard distance in the embedding space.

In deep metric learning, image features are extracted using a neural network $f: \mathbb{R}^{Height\times Width\times 3} \rightarrow \mathbb{R}^F$ producing an image representation vector $f(x)$, which is used as input for the encoder of the embedding $E: \mathbb{R}^F \rightarrow \mathbb{R}^D$. The latter is implemented as a fully connected layer generating an embedding vector of dimension $D$ used for computing similarities. The features $f$ and the encoder $E$ can then be trained jointly by standard back-propagation.

% Given an image representation $f(x)\in\mathbb{R}^F$, the encoder $E: \mathbb{R}^F \rightarrow \mathbb{R}^D$ outputs an encoding of dimension $D$ used for computing image similarities. 
With $d_{i j} = ||E(f(x_i)) - E(f(x_j))||^2$ defining the euclidean distance between the images $x_i$ and $x_j$, we require that $d_{i j} < d_{i k}$ if $y_j = y_i$ and $y_k \neq y_i$. 
% In order to train $f$ and $E$ simultaneously, the representation $f$ is implemented as a (convolutional) deep neural network and the encoder $E$ as a normalized single fully connected layer. We can then optimize them via stochastic gradient descent using standard triplet loss.
Given a triplet $(x_i,x_j,x_k)$ with $y_j = y_i$ and $y_k \neq y_i$, the loss is then defined as $l = \max{\left(d_{ij} - d_{ik} + m, 0\right)}$
% \begin{align}
%     l = max( d_{ij} - d_{ik} + m, 0)
% \end{align}
where $m$ is a margin parameter. Many variants of this loss have been proposed recently, with margin loss\cite{margin} (adding an additionally learnable margin $\beta$) proving to be best. \\

% \red{To encode discriminative and shared characteristics, we tackle the orthogonality of the two task by proposing a double encoder architecture where a class encoder $E_\alpha$ and an auxiliary encoder $E_\beta$ are trained simultaneously DIRECTLY EXPLAIN THE TWO ENCODING}(Fig.\ref{fig:model}). To efficiently train the deep neural network, the two encoders share the same image representation $f(x)$ which is updated by both tasks. The class encoder $E_\alpha$ is trained using the ground truth labels $y_1,\cdots,y_N$ and \red{any variant of the triplet loss EXPLAIN BETTER. A triplet loss variant} is also used to train the auxiliary encoder, however the labels are not provided. \red{Thus a surrogate task was developed to mine latent structured information from the data not affected by the ground truth classes.CHANGE}\\

\begin{table}
\begin{center}
\begin{tabular}{l|c|ccc|c}
\hline
R@k & Dim & 1 & 2 & 4 &  NMI \\
\hline
%LiftStruct \cite{lifted} & 512 & 46.6 & 58.1 & 69.8 & 56.2 \\
%FacilityLoc \cite{facilitylocation} & 64 & 48.2 & 61.4 & 71.8 & 59.2 \\
%SmartMin \cite{smartmining} & 64 & 49.8 & 62.3 & 74.1 & - \\
%Histogram\cite{histogram} & & 50.3 & 61.9 & 72.6  & - \\
% BinDev\cite{bindeviance} & - & 52.8 & 64.4 & 74.7  & - \\
% N-pairs\cite{npairs} & 64 & 51.0 & 63.3 & 74.3 &  60.4 \\
DVML\cite{dvml} & 512 & 52.7 & 65.1 & 75.5 &61.4 \\
% DAML\cite{daml} & 512 & 52.7 & 65.4 & 75.5 & 61.3 \\
%Angular\cite{angular} & 512 & 54.7 & 66.3 & 76.0& 61.1 \\
% HDC\cite{hdc} & 384 & 53.6 & 65.7 & 77.0 & - \\
BIER\cite{bier} & 512 & 55.3 & 67.2 & 76.9 & - \\
HTL\cite{htl} & 512 & 57.1 & 68.8 & 78.7 & - \\
A-BIER\cite{abier} & 512 & 57.5 & 68.7 & 78.3& - \\
HTG\cite{htg} & - & 59.5 & 71.8 & 81.3 & - \\
\hline
DREML\cite{dreml} & 9216 & 63.9 & 75.0 & 83.1 &  67.8 \\
\hline
Semihard\cite{semihard} & - & 42.6 & 55.0 & 66.4 & 55.4 \\
Semihard* & 128 & 57.2 & 69.4 & 79.9 & 63.9 \\
\textbf{MIC+semih} & 128 & 58.8 & 70.8 & 81.2 & 66.0 \\
ProxyNCA\cite{proxynca} & 64 & 49.2 & 61.9 & 67.9 & 64.9 \\
ProxyNCA* & 128 & 57.4 & 69.2 & 79.1 & 62.5 \\
\textbf{MIC+ProxyNCA} & 128 & 60.6 & 72.2 & 81.5 & 64.9 \\
Margin\cite{margin}     & 128 & 63.6 & 74.4 & 83.1 & 69.0 \\
Margin*                 & 128 & 62.9 & 74.1 & 82.9 &  66.3 \\
\textbf{MIC+margin}     & 128 & \textbf{66.1} & \textbf{76.8} & \textbf{85.6} & \textbf{69.7}\\
% \textbf{MIC+margin}     & 128 & \textbf{66.6} & \textbf{76.5} & \textbf{84.8} & \textbf{69.0} \\
% \hline
% Semihard* & 512 & 60.6 & 72.3 & 82.3 &  64.5 \\
% \textbf{COD+semih} & 512 & - & - & - & - \\
% % \textbf{COD+margin}  & 256 & - & - & - & - & - \\
% Margin* & 512 & 65.1 & 75.6 & 84.2 &  67.2 \\
% \textbf{COD+margin} & 512 & - & - & -  & - \\
% \hline\hline
\hline
\end{tabular}
\end{center}
\caption{Recall@k for k nearest neighbor and NMI on CUB200-2011 \cite{cub200-2011}. Our model outperforms all previous approaches, even those using a larger number of parameters. (*) indicates our best re-implementation with ResNet50.}
\label{tab:cub}
\end{table}

\subsection{Auxiliary Encoder}\label{sec:orthenc}
% We utilize two separate encodings to target the \blue{orthogonality} of finding both characteristics simultaneously. 
To separate the process of extracting both inter- and intra-class (shared) characteristics, we utilize two separate encodings: a class encoder $E_\alpha$ which aims to extract class-discriminative features and an auxiliary encoder $E_\beta$ to find shared properties. These encoders are trained together (Fig.\ref{fig:model}). To efficiently train the underlying deep neural network, the two encoders share the same image representation $f(x)$ which is updated by both during training. In the first training task, the class encoder $E_\alpha$ is trained using the provided ground truth labels $y_1,\cdots,y_N$ associated with each image $x_1,\cdots,x_N$ with $N$ the number of samples. A respective, metric-based loss function can be selected arbitrarily (such as a standard triplet loss or the aforementioned margin loss), as this part follows the generic training setup for metric learning problems. 
Because labels are not provided for the training of our auxiliary encoder, we define an automatic process to mine shared latent structure information from the original data. This information is then used to provide a new set of training labels to train our auxiliary encoder (Fig.\ref{fig:model} right). As the training scheme is now equivalent to the primary task, we may choose from the same set of loss functions.

\begin{table}
\begin{center}
\begin{tabular}{l|c|ccc|c}
\hline
R@k & Dim & 1 & 2 & 4  & NMI \\
\hline
%LiftStruct \cite{lifted} & 512 & 48.3 & 61.1 & 71.8 & 55.1 \\
%FacilityLoc \cite{facilitylocation} & 64 & 67.0 & 83.7 & 93.2 & - \\
%SmartMin \cite{smartmining} & 64 & 64.7 & 76.2 & 84.2 & - \\
% N-pairs\cite{npairs} & 64 & 71.1 & 79.7 & 86.5 & - \\
%Angular\cite{angular} & 512 & 71.4 & 81.4 & 87.5& - \\
% HDC\cite{hdc} & 384 & 73.7 & 83.2 & 89.5 & - \\
% DAML\cite{daml} & 512 & 75.1 & 83.8 & 89.7 & 66.0 \\
HTG\cite{htg} & - & 76.5 & 84.7 & 90.4 & - \\
BIER\cite{bier} & 512 & 78.0 & 85.8 & 91.1& - \\
HTL\cite{htl} & 512 & 81.4 & 88.0 & 92.7  & - \\
DVML\cite{dvml} & 512 & 82.0 & 88.4 & 93.3 & 67.6 \\
A-BIER\cite{abier} & 512 & 82.0 & 89.0 & 93.2  & - \\
\hline
DREML\cite{dreml} & 9216 & 86.0 & 91.7 & 95.0 & 76.4 \\
\hline
Semihard\cite{semihard} & - & 51.5 & 63.8 & 73.5 & 53.4 \\
Semihard* & 128 & 65.5 & 76.9 & 85.2 & 58.3 \\
\textbf{MIC+semih} & 128 & 70.5 & 80.5 & 87.4 & 61.6 \\
ProxyNCA\cite{proxynca} & 64 & 73.2 & 82.4 & 86.4 & - \\
ProxyNCA* & 128 & 73.0 & 81.3 & 87.9 & 59.5 \\
\textbf{MIC+ProxyNCA} & 128 & 75.9 & 84.1 & 90.1 & 60.5 \\
Margin\cite{margin}     & 128 & 79.6 & 86.5 & 90.1 & \textbf{69.1} \\
Margin*                 & 128 & 80.0 & 87.7 & 92.3 & 66.3 \\
\textbf{MIC+margin}     & 128 & \textbf{82.6} & \textbf{89.1} & \textbf{93.2} & 68.4\\
% \textbf{MIC+margin}     & 128 & \textbf{84.1} & \textbf{90.5} & \textbf{94.2} & \textbf{70.0} \\
% \hline
% Semihard* & 512 & 75.1 & 83.6 & 89.8 & 62.8\\
% \textbf{COD+semih} & 512 & - & - & - & - \\
% \textbf{COD+margin}  & 256 & 81.1 & 88.2 & 92.6 & 96.0 & 68.8 \\
% Margin* & 512 & 80.1 & 87.1 & 91.7 & 64.1 \\
% \textbf{COD+margin} & 512 & - & - & - & - \\
% \hline\hline
\hline
\end{tabular}
\end{center}
\caption{Recall@k for k nearest neighbor and NMI on CARS196 \cite{cars196}. DREML\cite{dreml} is not comparable given the large embedding dimension. (*) indicates our ResNet50 re-implementation.}
\label{tab:cars}
\end{table}

\subsection{Extracting Inter-class Characteristics}\label{sec:surrogate}
We seek a task which, without human supervision, spots structured characteristics within the data while ignoring class-specific information. 
% Structured properties are defined by characteristics shared among several images, therefore creating homogeneous groups.
% \blue{To spot structured characteristics within the data void of class-specific information, we seek a training task without additional human supervision. }
As structured properties are generally defined by characteristics shared among several images, they create homogeneous groups. To find these, clustering offers a well established solution.
This algorithm associates images to surrogate labels $c_1,\cdots,c_N$ with $c_i \in [1,\cdots,C]$ and $C$ being the predefined number of clusters. 
However, applied directly to the data, this method is biased towards class-specific structures since images from the same class share many common properties, like color, context and shape, mainly injected through the data collection process (e.g. a class may be composed of pictures of the same object from multiple angles).

To remove the characteristics shared within the class, %(e.g. specific color schemes or orientations), 
we apply normalization guided by the ground truth classes. For each class $y$ we compute the mean $\mu_y$ % = \frac{1}{N_y} \sum_{i: y_i==y} f(x_i)$ 
and standard deviation $\sigma_y$ based on the features $f(x_i),  \forall x_i: y_i=y$. Then we obtain the new standardized image representation $Z = [z_1,\cdots,z_N]$%$z_i, i\in [1,\cdots,N]$ 
with $z_i = \frac{f(x_i) - \mu_{y_i}}{\sigma_{y_i}}$, where the class influence is now reduced. 
Afterwards, the auxiliary encoder $E_\beta$ can be trained using the surrogate labels $[c_1,\cdots,c_N]$ produced by clustering the space $Z$. 

% \red{For that to be successful however, it is [crucial] SOUNDS LIKE A WEAKNESS that for the initial label extraction a strong prior is given WE NEED TO MENTION THAT F(X) IS INITIALIZED WITH IMAGENET which we can reinforce through training $E_\beta$. See also Caron \etal.\cite{deepcluster} who use a similar setup to train an unsupervised ImageNet classifier. SOUNDS LIKE WE ARE COPYING THEM. ALSO, IS THIS THE BEST POSITION? MAYBE BETTER IN RELATED WORK}
For that to work as intended, a strong prior is needed. It is standard procedure for deep metric learning to initialize the representation backend $f$ with weights pretrained on ImageNet. This provides a sufficiently good starting point for clustering, which is then reinforced through training $E_\beta$.

% The surrogate task and the encoder training are summarized in Fig.\ref{fig:model}. Fig.\ref{fig:clusters} shows some examples of clusters detected using our surrogate task.
Fig.\ref{fig:clusters} shows some examples of clusters detected using our surrogate task. This task and the encoder training are summarized in Fig.\ref{fig:model}.

\begin{table}
\begin{center}
\begin{tabular}{l|c|ccc|c}
\hline
R@k & Dim & 1 & 10 & 100 & NMI \\
\hline
%Histogram\cite{histogram} & - & 63.9 & 81.7 & 92.2 &  - \\
% BinDev\cite{bindeviance} & - & 65.5 & 82.3 & 92.3 &  -\\
%LiftStruct \cite{lifted} & 512 & 63.0 & 80.5 & 91.7 & 87.4 \\
%FacilityLoc \cite{facilitylocation} & 64 & 67.0 & 83.7 & 93.2 & -  \\
% N-pairs\cite{npairs} & 512 & 67.7& 83.8 & 93.0 & 88.1 \\
%Angular\cite{angular} & 512 & 70.9 & 85.0 & 93.5 & 88.6 \\
% DAML\cite{daml} & 512 & 68.4 & 83.5 & 92.3 & 89.4 \\
% HDC\cite{hdc} & 384 & 69.5 & 84.4 & 92.8 & - \\
DVML\cite{dvml} & 512 & 70.2 & 85.2 & 93.8 & \textbf{90.8} \\
BIER\cite{bier} & 512 & 72.7 & 86.5 & 94.0 & - \\
ProxyNCA\cite{proxynca} & 64 & 73.7 & - & - & - \\
A-BIER\cite{abier} & 512 & 74.2 & 86.9 & 94.0 & - \\
% HTL\cite{htl} & 512 & 69.5 & 84.4 & 92.8 &  - \\
HTL\cite{htl} & 512 & 74.8 & 88.3 & 94.8 &  - \\
\hline
%Semi-hard\cite{semihard} & - & 66.7 & 82.4 & 91.9 &  89.5 \\
%Semihard* & 128 & 67.4 & 83.9 & 92.3 & 88.5 \\
%\textbf{COD+semih} & 128 & - & - & - & - \\
Margin\cite{margin}     & 128 & 72.7 & 86.2 & 93.8 & \textbf{90.7} \\
Margin*                 & 128 & 74.4 & 87.2 & 94.0 & 89.4 \\
\textbf{MIC+margin}    & 128 & \textbf{77.2} & \textbf{89.4} & \textbf{95.6} &  90.0\\
% \textbf{MIC+margin}     & 128 & \textbf{76.1} & \textbf{88.1} & \textbf{94.8} &  89.9 \\
% \hline
% \textbf{MIC**+margin}   & 128 & \textbf{76.9} & 90.0 & \textbf{96.4} & 90.1 \\
% \hline
% Semihard* & 512 & 70.1 & 85.7 & 92.7 & 88.6 \\
% \textbf{COD+semih} & 512 & - & - & - & - \\
% % \textbf{COD+margin}  & 256 & - & - & - &  - \\
% Margin*                 & 512 & 73.3 & 87.8 & 95.1 & \textbf{91.0} \\
% \textbf{COD+margin}     & 512 & - & - & - &  - \\
\hline
\end{tabular}
\end{center}
\caption{Recall@k for k nearest neighbor and NMI on Stanford Online Products \cite{lifted}. (*) indicates our ResNet50 re-implementation.}
% (**) denotes training of $E_\beta$ with low-grained object categories.
\label{tab:sop}
\end{table}

\begin{table}
\begin{center}
\begin{tabular}{l|c|cccc}
\hline
R@k & Dim & 1 & 10 & 30 & 50 \\
\hline
% FashionNet\cite{inshop}    & -   & 53.0 & 73.0 & 77.0 & 80.0 \\
% HDC\cite{hdc}              & 384 & 62.1 & 84.9 & 91.2 & 93.1 \\
BIER\cite{bier}            & 512 & 76.9 & 92.8 & 96.2 & 97.1 \\
HTG\cite{htg}              & -   & 80.3 & 93.9 & 96.6 & 97.1 \\
HTL\cite{htl}              & 512 & 80.9 & 94.3 & 97.2 & 97.8 \\
A-BIER\cite{abier}         & 512 & 83.1 & 95.1 & 97.5 & 98.0 \\
\hline
DREML\cite{dreml}          & 9216& 78.4 & 93.7 & 96.7 & - \\
\hline
Margin*             & 128 & 84.5 & 95.7 & 97.6 & 98.3 \\
\textbf{MIC+margin} & 128 & \textbf{88.2} & \textbf{97.0} & \textbf{98.0} & \textbf{98.8}\\
% \textbf{MIC+margin} & 128 & \textbf{87.1} & \textbf{97.0} & \textbf{98.2} & \textbf{98.6} \\
\hline
\end{tabular}
\end{center}
\caption{Recall@k for k nearest neighbor and NMI on In-Shop \cite{inshop}. (*) indicates our best re-implementation with ResNet50}
\label{tab:inshop}
\end{table}

\begin{table}
\begin{center}
\begin{tabular}{l|c|cc|cc}
\hline
\multicolumn{2}{l}{Test Splits} & \multicolumn{2}{c}{Small} & \multicolumn{2}{c}{Large} \\
\hline
R@k & Dim & 1 & 5 & 1 & 5 \\
\hline
MixDiff+CCL\cite{pku}   & -   & 49.0 & 73.5 & 38.2 & 61.6 \\
GS-TRS\cite{gstrs}    & -   & 75.0 & 83.0 & 73.2 & 81.9 \\
BIER\cite{bier}            & 512 & 82.6 & 90.6 & 76.0 & 86.4 \\
A-BIER\cite{abier}         & 512 & 86.3 & 92.7 & 81.9 & 88.7 \\
\hline
DREML\cite{dreml}          & 9216& 88.5 & 94.8 & 83.1 & 92.4 \\
\hline
Margin*             & 128 & 85.1 & 92.4 & 80.4 & 88.9 \\
\textbf{MIC+margin} & 128 & \textbf{86.9} & \textbf{93.4} & \textbf{82.0} & \textbf{91.0} \\
\hline
\end{tabular}
\end{center}
\caption{Recall@k for k nearest neighbor and NMI on PKU VehicleID\cite{pku}. DREML\cite{dreml} is not comparable given the large embedding dimension. (*) our best ResNet50 re-implementation}
\label{tab:pku}
\end{table}

\begin{figure}[t]
\begin{center}
   \includegraphics[width=0.88\linewidth]{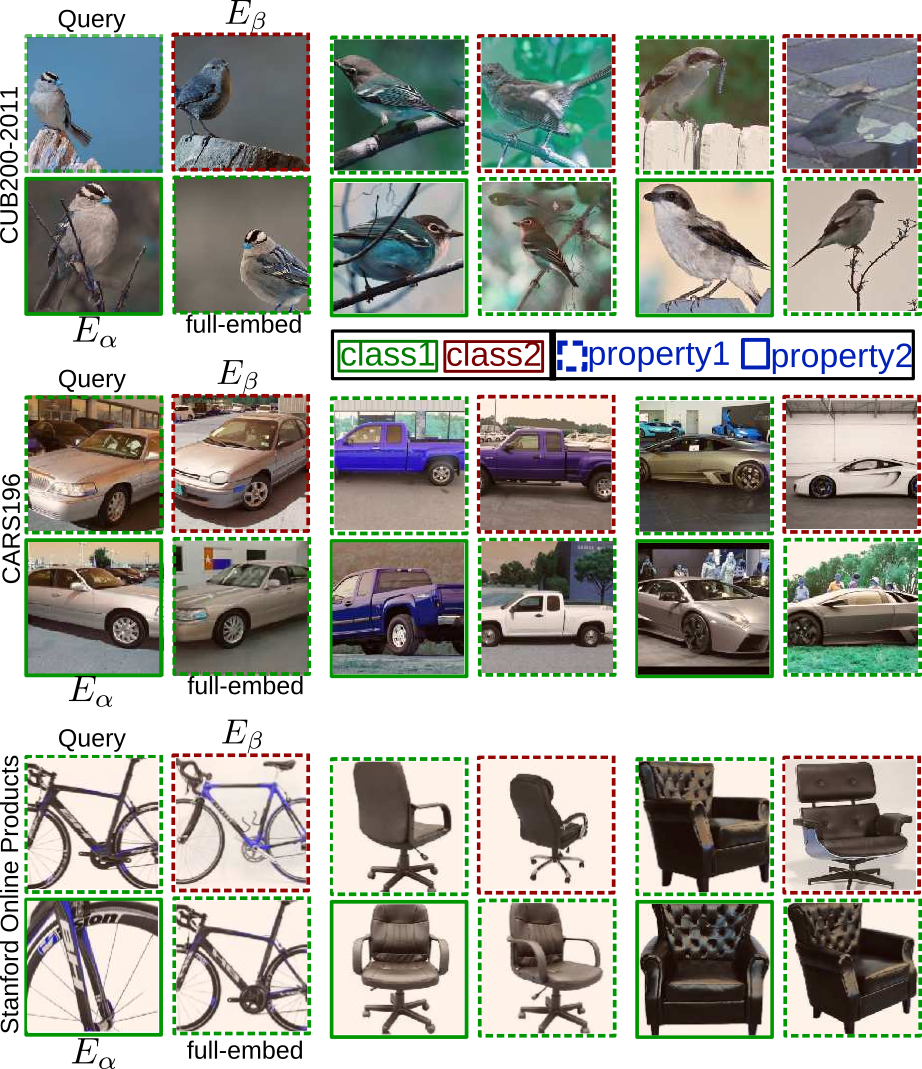}
\end{center}
   \caption{Qualitative nearest neighbor evaluation for CUB200-2011, CARS196 and SOP based on $E_\alpha$ and $E_\beta$ encodings and their combination. The results show that $E_\beta$ leverages class-independent information (posture,parts) while $E_\alpha$ becomes independent to those features and focuses on the class detection. The combination of the two reintroduces both.}% that $E_\alpha$.}% needed to rid itself of.}
\label{fig:nn}
\end{figure}

% \subsection{Minimize Embeddings Mutual Information}\label{sec:loss}
\subsection{Minimizing Mutual Information}\label{sec:loss}

The class encoder $E_\alpha$ and auxiliary encoder $E_\beta$ can then be trained using the respective labels. As we utilize two different learning tasks,  $E_\alpha$ and $E_\beta$ learn distinct characteristics. However, as both share the same input, the image features $f(x)$, a dependency between the encoders can be induced, therefore leading to both encoders learning some similar properties. To reduce this effect and to constrain the discriminative and shared characteristics into their respective encoding space, we introduce a mutual information loss, which we compute through an adversarial setup
% Formally, we do so by maximizing the euclidean distance $l_d = || E_\alpha(f(x)) - E_\beta(f(x)) ||^2$ between $E_\alpha$ and $E_\beta$ for the image $x$.
% We target this by maximizing the forward Kullback Leibler Divergence. In our regressive problem, the forward KL Divergence translates to Maximum Likehood Estimation \cite{kldiv2mle}. Under a normal data distribution, this allows us to formulate this as maximization of the euclidean distance $l_d = || E_\alpha(f(x)) - E_\beta(f(x)) ||^2$ between same-sized $E_\alpha$ and $E_\beta$ for the image $x$.
% \blue{We target this by adding an adversarial loss similar to the proposal in \cite{abier} to ensure high dissimilarity between $E_\alpha$ and $E_\beta$, which we define as 
% \begin{equation}
%     l_d = \left( E_\alpha(f(x)) \odot R(E_\beta(f(x))) \right)^2
% \end{equation}
% with $R$ being a two-layered fully-connected neural network with normalized output projecting $E_\beta$ to the encoding space of $E_\alpha$. Because we enforce $E_\alpha$, $E_\beta$ and $R(E_\beta)$ to have unit length, we do not perform additional regularization like \cite{abier}.
% Using this setup, we are not limited in the dimensionalities used for $E_\beta$, which becomes relevant in our ablation studies (see section \ref{sec:embprop}).}
%\blue{%We target this through a mutual information loss
% Inspired by \cite{abier}, we minimize the mutual information using
\begin{equation}
    l_d = -\left( E^r_\alpha(f(x)) \text{\large{$\odot$}} R(E^r_\beta(f(x))) \right)^2
\end{equation}
with $R$ being a learned, small two-layered fully-connected neural network with normalized output projecting $E_\beta$ to the encoding space of $E_\alpha$. 
% The $r$ superscript ensures the adversarial setup through an appended gradient reversal layer \cite{gradreverse} s.t. when trying to minimize $l_d$ the similarity between both encoders actually decreases.
$\odot$ stands for an elementwise product, while the $r$ superscript notes a gradient reversal layer \cite{gradreverse} which flips the gradient sign s.t. when trying to minimize $l_d$, i.e. maximizing correlation, the similarity between both encoders is actually decreased.
% Since we enforce $E_\alpha$, $E_\beta$ and $R(E_\beta)$ to have unit length, we do not perform additional regularization.
A similar method has been adopted by \cite{abier}, %where an ensemble of encoders are trained by minimizing their shared information. 
where shared information is minimized between an ensemble of encoders.
In contrast, our goal is to transfer non-discriminate characteristics to an auxiliary encoder. Finally, as $l_d$ scales with $R$, we avoid trivial solutions (e.g. $R(E_\beta) \rightarrow \infty$) by enforcing $R(E_\beta)$ to have unit length, similar to $E_\alpha$ and $E_\beta$.
%like \cite{abier}.
% Using this setup, we are not limited in the dimensionalities used for $E_\beta$, which becomes relevant in our ablation studies (see section \ref{sec:embprop}).}

% \begin{align}
%     l_d &= || E_\alpha(f(x)) - E_\beta(f(x)) ||^2.
% \end{align}

Finally, the total loss $L$ to train our two encoders and the representation $f$ is computed by $L = l_\alpha + l_\beta + \gamma l_d$, 
% \begin{align}\label{eq:loss}
%     L &= l_\alpha + l_\beta - \gamma l_d
% \end{align}
where $\gamma$ weights the contribution of the mutual information loss with respect to the class triplet loss $l_\alpha$ and the auxiliary triplet loss $l_\beta$.
The full training is described in Alg. \ref{alg:algorithm}.

%-------------------------------------------------------------------------

\section{Experiments}
In this section we offer a quantitative and qualitative analysis of our method, also in comparison to previous work. After providing technical information for reproducing the results of our model, we give some information regarding the standard benchmarks for metric learning and provide comparisons to previous methods. Finally, we offer insights into the model by studying its key components.

\begin{figure}[t]
\begin{center}
   \includegraphics[width=0.95\linewidth]{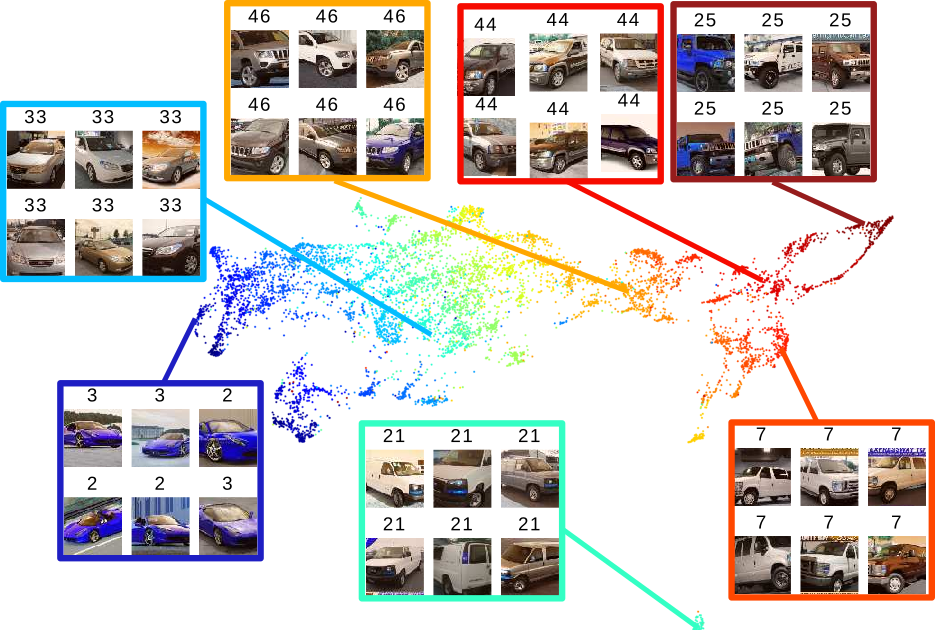}
\end{center}
%   \caption{UMAP projection of the $E_\alpha$ encoding for CARS196. The data is clustered to groups of images considered similar by the encoder. The cluster ids are sorted by the y-axis for visual appeal. We selected seven clusters throughout the full distribution and show the images close to the centroid. The ground-truth label is shown above the image. The result shows that the encoding extracts class-specific information and ignores others (posture, etc).}
   \caption{UMAP projection of $E_\alpha$ for CARS196. Seven clusters are selected, showing six images near the centroid and their ground-truth labels. We see that the encoding extracts class-specific information and ignores other (e.g. orientation).}
\label{fig:classumap}
\end{figure}

\subsection{Implementation details}\label{sec:impl}
We implement our method using the PyTorch framework \cite{pytorch}. 
% We conducted all the experiments using ResNet50\cite{resnet}, as this architecture is most widely used in recent metric learning work. 
As baseline architecture, we utilize ResNet50 \cite{resnet} due to its widespread use in recent metric learning work.
All experiments use a single NVIDIA GeForce Titan X.
% Practically, we train the main and surrogate task alternating from one to another, one iteration each, maximizing the batch-size on a single GPU.
Practically, class and auxiliary encoders $E_\alpha$ and $E_\beta$ use the same training protocol (following \cite{margin} with embedding dimensions of $128$) with alternating iterations to maximize the usable batch-size.
% \cite{margin} propose a dimensionality of $128$ for the class encoder $E_\alpha$ used for the evaluations. 
The dimensionality of the auxiliary encoder $E_\beta$ is fixed (except for ablations in sec. \ref{sec:ablations}) to the dimensionality of $E_\alpha$ to ensure similar computational efficiency compared to previous work.
However, due to GPU memory limitations, we use a batchsize of $112$ instead of a proposed $128$, with no relevant changes in performance.
% In fact, since the clustering has insignificant computational cost, the additional computation with respect to \cite{semihard,margin} comes from the size of $E_\beta$.
% As the clustering step has minor computational cost (additional overhead of roughly $10-15\%$, the additional computation with respect to \cite{semihard,margin} comes mainly from the size of $E_\beta$.
% Since the back-propagation of the surrogate task is parallel to the main task, i.e. back-propagating both count as two epochs, the only computational cost comes from computing the clusters, which has insignificant cost respect to the back-propagation. our approach requires only an additional $80\%$ in time complexity respect to \cite{semihard,margin} due to the back-propagation of the surrogate task, which is one or two order of magnitude less than other ensemble methods\cite{dreml,abier,bier} competitive with our method.

% During training, we randomly crop images of size $224 \times 224$ after resizing to $256 \times 256$, and randomly flip the image horizontally. 
During training, we randomly crop images of size $224 \times 224$ after resizing to $256 \times 256$, followed by random horizontal flips.
For all experiments, we use the original images without bounding boxes. We train the model using Adam \cite{adam} with a learning rate of $10^{-5}$ and set the other parameters to default. %For all datasets, we use the mentioned batch size of $112$. 
We set the triplet parameters following \cite{margin}, initializing $\beta = 1.2$ for the margin loss and $\alpha = 0.2$ as fixed triplet margin. Per mini-batch, we sample $m = 4$ images per class for a random set of classes, until the batch size is reached.
For $\gamma$ (Sec. \ref{sec:loss} eq.) we utilize dataset-dependent values in $[100,2000]$ determined via cross-validation.
%suggested in \cite{abier}.

After class standardization, the clustering is performed via standard k-means using the faiss framework \cite{faiss}. 
%\red{Added computation time scales with the size of the dataset and the dimension of $E_\beta$. Using our suggested hyperparameters (see below), the overhead is between $10-25\%$ of total training time, depending on whether the cluster computation is performed on a GPU or a CPU}. 
Using the hyperparameters proposed in this paragraph, the computational cost introduced by our approach is 10-20\% of total training time. For efficiency, the clustering can be computed on GPU using faiss\cite{faiss}.
The number of clusters is set before training to a fixed, problem-specific value: $30$ for CUB200-2011 \cite{cub200-2011}, $200$ for CARS196 \cite{cars196}, $50$ for Stanford Online Products \cite{lifted}, $150$ for In-Shop Clothes \cite{inshop} and $50$ for PKU VehicleID \cite{pku}.
%share the same cluster value with Stanford Online Products due to a similar dataset makeup (lots of classes, small number of images per class).
We update the cluster labels every other epoch.
Notably, however, our model is robust to both parameters since a large range of parameters give comparable results. Later in section \ref{sec:ablations} we study the effect of cluster numbers and cluster label update frequencies for each dataset in more detail to motivate the chosen numbers. 
%\red{Finally, during training of the auxiliary embedding, we randomly switch positive/anchor samples with samples from different cluster classes with probability $p=0.2$, as we have found this to improve convergence.}
Finally, class assignments by clustering, especially in the initial training stages, becomes near arbitrary for samples further away from cluster centers. To ensure that we do not reinforce such a strong initial bias, we found it beneficial to ease the class constraint by randomly switching samples with samples from different cluster classes (with probability $p\leq 0.2$).

\begin{figure}[t]
\begin{center}
   \includegraphics[width=0.90\linewidth]{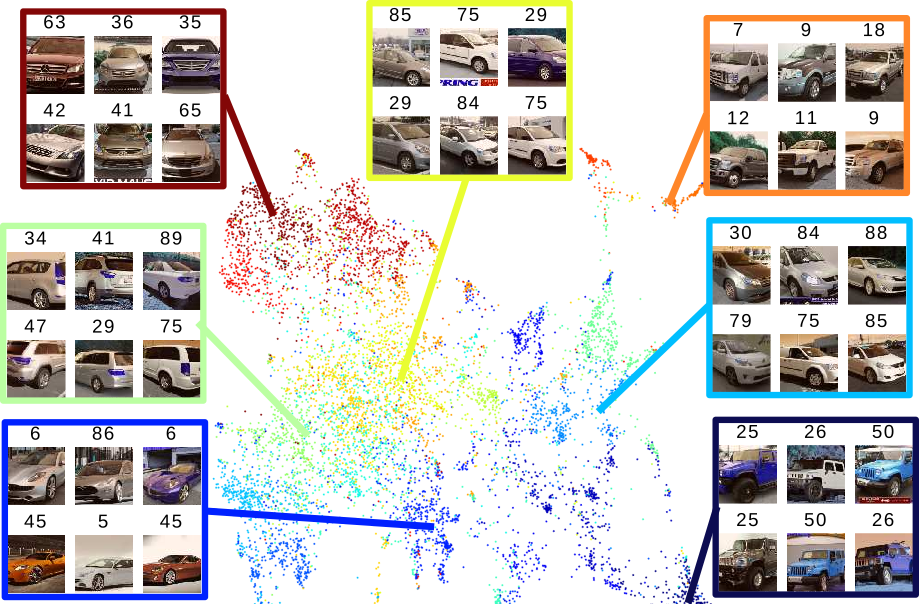}
\end{center}
%   \caption{UMAP projection of the $E_\beta$ encoding for CARS196. The data is clustered into groups of images considered similar by the encoder. The cluster ids are sorted by the y-axis for visual appeal. We selected seven clusters throughout the full distribution and show the images close to the centroid. The ground-truth label is shown above the image. The result shows that the encoding extracts intrinsic characteristics of the object (car) independent of the classes label.}
   \caption{UMAP projection of $E_\beta$ for CARS196. Seven clusters are selected, showing six images near the centroid and their GT labels. The result shows that the encoding extracts intrinsic characteristics of the object (car) independent from GT classes.}

\label{fig:attributeumap}
\end{figure}

\subsection{Datasets}
Our model is evaluated on five standard benchmarks for image retrieval typically used in deep metric learning. We report the Recall@k metric \cite{recall} to evaluate image retrieval and the normalized mutual information score (NMI) \cite{nmi} for the clustering quality. The training and evaluation procedure follows the standard setup as used in \cite{margin}.\\
% \textbf{CARS196\cite{cars196}: }offers 196 different car models, totaling 16,185 images. 
% We use the first 98 classes($8054$ images) for training and the remaining 98 for testing ($8131$ images).\\
\textbf{CARS196\cite{cars196} }with 196 car models over 16,185 images. 
We use the first 98 classes ($8054$ images) for training and the remaining 98 ($8131$ images) for testing.\\
% \textbf{Stanford Online Products\cite{lifted}: }contains 22,634 object classes in 120,053 images from 12 product categories. Following \cite{margin}, 59,551 images (11,318 classes) are used during train and 60,502 (11,316 classes) for evaluation. \\
\textbf{Stanford Online Products\cite{lifted} }with 120,053 product images in 22,634 classes. 59,551 images (11,318 classes) are used for training, 60,502 (11,316 classes) for testing. \\
% \textbf{CUB200-2011\cite{cub200-2011}: }provides a collection of 200 bird species for a total of 11,788 images. As before, the first half, here 100 classes (5,864 images), is used for training and the remaining 100 classes (5,924 images) for evaluation.\\
\textbf{CUB200-2011\cite{cub200-2011} }with 200 bird species over 11,788 images. Train and Test Sets contain the first and last 100 classes (5,864/5,924 images) respectively.\\
% \textbf{In-Shop Clothes\cite{inshop}: }offers a collection of clothing images in 7,986 classes over 52,712 images. Following \cite{inshop}, we use a training subset containing 3,997 classes and a test subset with 3,985 classes. Additionally, the test set is divided into a query set with 14,218 images and a gallery set with 12,612 for evaluation.\\
\textbf{In-Shop Clothes\cite{inshop} }with 72,712 clothing images in 7,986 classes. 3,997 classes are used for training and 3,985 classes for evaluation. The test set is divided into a query set (14,218 images) and a gallery set (12,612 images).\\
% \textbf{PKU VehicleID\cite{pku}: }contains 221,736 vehicle images captured by surveillance cameras, distributed over 26,267 vehicles. This set comprises the training set with 110,178 images over 13,134 classes and the testing set with 111,585 images of 13,133 vehicles. With less intra-class variance compared to CARS196, the retrieval difficulty in this dataset stems from different vehicle identities for the same car models.
% We follow the training and evaluation instructions provided by \cite{pku} and use predefined testing subsets with 7,332 (small) and 20,038 (large) images respectively. For tabular clarity, we ignore the medium test set. \\
% \textbf{PKU VehicleID\cite{pku}: }contains 221,736 surveillance images of 26,267 vehicles with shared car models. We follow the training and evaluation instructions provided by \cite{pku} and use 110,178 images over 13,134 classes for training. Testing is done on predefined testing subsets with 7,332 (small) and 20,038 (large) images respectively. For tabular clarity, we disregard the medium test set.\\
\textbf{PKU VehicleID\cite{pku} }with 221,736 surveillance images of 26,267 vehicles with shared car models. We follow \cite{pku} and use 13,134 classes (110,178 images) for training. Testing is done on a predefined small and large testing subset with 7,332 (small) and 20,038 (large) images respectively.\\

\begin{figure}[t]
\begin{center}
   \includegraphics[width=0.90\linewidth]{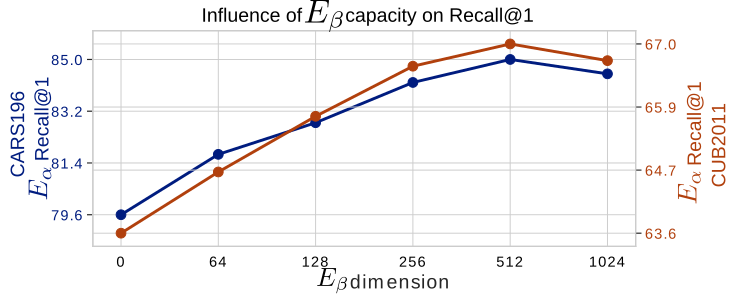}
\end{center}
%   \caption{Evaluation of $E_\alpha$ as a function of the $E_\beta$ capacity. The y-axis represents the Recall@1 using $E_\alpha$ for inference on CARS196\cite{cars196} and CUB200-2011\cite{cub200-2011}, and the x-axis the dimensionality of $E_\beta$ during training. The results show that increasing the capacity of the auxiliary encoder $E_\beta$, thus being able to better learn the properties shared among classes, directly benefits the class encoder $E_\alpha$.}
   \caption{Evaluation of $E_\alpha$ as a function of the $E_\beta$ capacity. For CARS196 \cite{cars196} and CUB200-2011 \cite{cub200-2011}, we plot $E_\alpha$ Recall@1 against the $E_\beta$ dimension during training. The results show that the increase in capacity of $E_\beta$ and thus the ability to learn properties shared among classes directly benefits the class encoder $E_\alpha$.}
\label{fig:recall_betadim}
\end{figure}

\subsection{Quantitative and Qualitative Results}
In this section we compare our approach with existing models from recent literature. Our method is applied on three different losses, the standard triplet loss with semi-hard negative mining \cite{semihard}, Proxy-NCA \cite{proxynca} and the state-of-the-art margin loss with weighted sampling \cite{margin}. For full transparency, we also provide results with our re-implementation of the baselines.% of each method.% in a single framework.

The results show a consistent gain over the state of the art for all datasets% in proportion to available class variance
, see tables \ref{tab:cub}, \ref{tab:cars}, \ref{tab:sop}, \ref{tab:inshop} and \ref{tab:pku}. In particular, our approach achieves better results than more complex ensembles. On CUB200-2011, we outperform even DREML \cite{dreml} which trains $48$ ResNet models in parallel. %In addition, we see the performance gain independently of the loss used, see Tab.\ref{tab:cub} and \ref{tab:cars}.

Qualitative results are shown in Fig.\ref{fig:nn}: the class encoder $E_\alpha$ retrieves images sharing class-specific characteristics, while the auxiliary encoder $E_\beta$ finds intrinsic, class-independent object properties (e.g. posture, context). The combination retrieves images with both characteristics.

\section{Ablations}\label{sec:ablations}
In this section, we investigate the properties of our model and evaluate its components. We qualitatively examine the proposed encoder properties by checking recalled images for both and study the influence of $E_\beta$ on the recall performance, see Section \ref{sec:embprop}. In Section \ref{sec:ablations} we measure the relation between the intra-class variance and the capacity of our auxiliary encoder $E_\beta$. In addition, ablation studies are performed to examine the relevance of each pipeline component and hyper-parameter. We primarily utilize the most common benchmarks CUB200-2011, CARS196 and SOP.

\begin{figure}[t]
\begin{center}
   \includegraphics[width=0.90\linewidth]{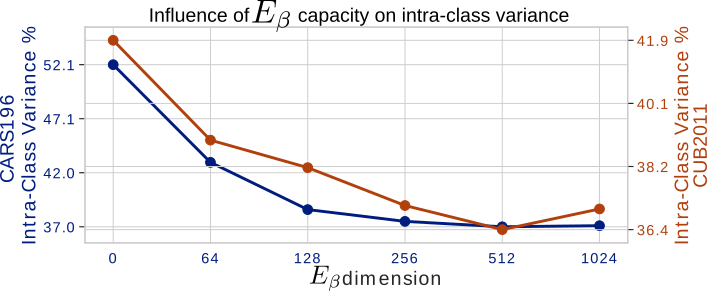}
\end{center}
   \caption{Measure of the intra-class variance in the class embedding $E_\alpha$ as function of the auxiliary encoder $E_\beta$ dimension. The result shows that the intra-class variance decreases with an increase in $E_\beta$ capacity. This points towards $E_\beta$ making it easier for $E_\alpha$ to disregard class-independent information.}
\label{fig:intravariance_betadim}
\end{figure}

\subsection{Embedding Properties}\label{sec:embprop}
Firstly, we visualize the characteristics of the class encoder $E_\alpha$ (Fig.\ref{fig:classumap}) and auxiliary encoder $E_\beta$ (Fig.\ref{fig:attributeumap}) by projecting the embedded test data to two dimensions using UMAP\cite{umap}. The figures show $E_\alpha$ extracting class-discriminative information while $E_\beta$ encodes characteristics shared across classes (e.g. car orientation).

To evaluate the effect of the auxiliary encoder $E_\beta$ on the class encoder $E_\alpha$, we study the properties of the class encoding as function of the capability of $E_\beta$ to learn shared characteristics. 
% First, we study the performance of $E_\alpha$ on CARS196\cite{cars196}, CUB200-2011\cite{cub200-2011} and SOP\cite{lifted} when changing the dimensionality of the auxiliary encoder.
First, we study the performance of $E_\alpha$ on CARS196\cite{cars196} and  CUB200-2011\cite{cub200-2011} relative to the auxiliary encoder dimension. Utilizing varying $E_\beta$ dimensionalities, Fig.\ref{fig:recall_betadim} shows a direct relation between $E_\beta$ capacity and the retrieval capability. $E_\beta$ with dimension $0$ indicates the baseline method \cite{margin}.
% For all other evaluations we fix the $E_\beta$ dimensionality equal to the dimension of $E_\alpha$ to keep the computational complexity comparable to baseline\cite{margin} (see Sec.\ref{sec:impl}).
For all other evaluations, the $E_\beta$ dimension is equal to $E_\alpha$ to keep the computational cost comparable to the baseline \cite{margin} (see Sec.\ref{sec:impl}).

% In addition, we assume 
% Our main intuition behind this paper is that explicitly learning the shared latent characteristics, instead of becoming invariant to them, is a better solution for more compact class representations. 
% In addition, we assume that explicitly learning the shared latent characteristics, instead of becoming invariant to them, is a better solution for more compact class representations. 

% Our initial assumption is that learning shared characteristics produces more compact classes.
% We study the intra-class variance by computing 
To examine our initial assumption that learning shared characteristics produces more compact classes, we study the intra-class variance by computing 
% To examine that, we look at the intra-class variance of the produced embedding by looking at 
the mean pairwise distances per class, averaged over all classes. These distances are normalized by the average inter-class distance, approximated by the distance between two class centers.%, for comparability between different encodings.
Summarized in fig.\ref{fig:intravariance_betadim} we see higher intra-class variance for basic margin loss ($E_\beta$ dimension equal to $0$).
% More importantly, the capacity of $E_\alpha$ to produce compact classes is directly related to the capacity of the auxiliary encoder $E_\beta$. 
But more importantly, the class compactness is directly related to the capacity of the auxiliary encoder $E_\beta$. 

We also offer a qualitative evaluation of the surrogate task in Fig.\ref{fig:clusters}. % shows some qualitative results of our surrogate task. 
After class-standardization, the clustering recognizes latent structures of the data shared across classes.

\begin{figure}[t]
\begin{center}
   \includegraphics[width=0.90\linewidth]{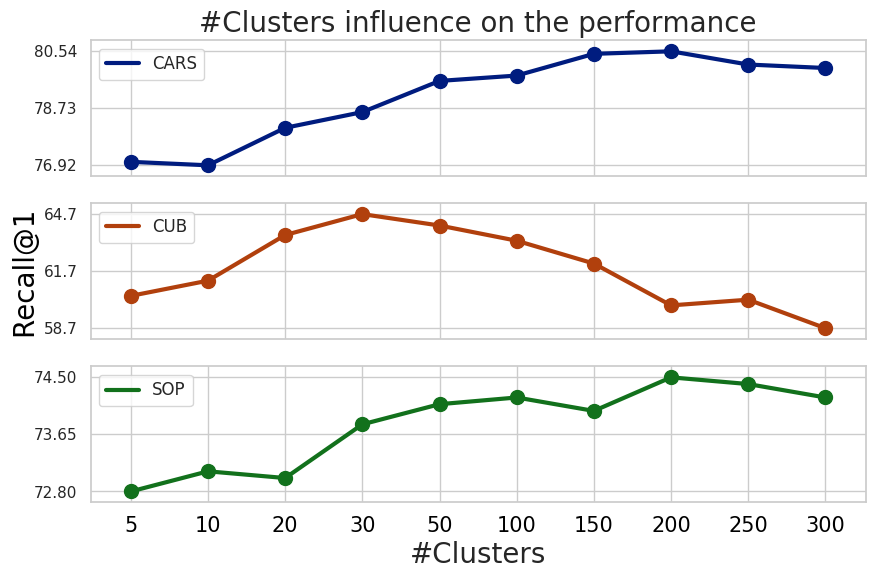}
\end{center}
   \caption{Ablation study: influence of the number of clusters on Recall@1. A fixed cluster label update period of 1 was used with equal learning rate and consistent scheduling.}
\label{fig:cluster_recall}
\end{figure}

% \begin{table}
% \begin{center}
% \begin{tabular}{l|c|c|c}
% \hline
% Dataset & CARS196 & CUB200 & SOP \\
% \hline\hline
% OurBaseline\cite{margin} & 80.0 & 62.9 & 73.2 \\
% \hline
% +Kmean,-Standard & 79.2 & 59.1 & 71.9 \\
% eval on $E_\beta$ & 79.6 & 60.2 & 66.2 \\
% NoMutualInfo($\gamma=0$)  & 82.6 & 64.1 & 74.9 \\
% FullModel         & \textbf{84.0} & \textbf{66.6} & \textbf{76.1} \\
% \hline
% \end{tabular}
% \end{center}
% \caption{Ablation study: Relevance of different contributions. Each component is crucial for reaching the best performances, }
% \label{tab:components}
% \end{table}

\begin{table}
\begin{center}
\begin{tabular}{c|c|c||c|c|c}
\hline
Clust & Stand & MutInfo & CARS & CUB & SOP \\
\hline
- & - & - & 80.0 & 62.9 & 73.2 \\
+ & - & - & 79.2 & 59.1 & 71.9 \\
% + & + & - & 83.1 & 65.3 & 74.9 \\ Wrong SOP value
% + & + & - & 83.1 & 65.3 & 75.8 \\
+ & + & - & 81.3 & 64.9 & 75.8 \\
% + & + & + & \textbf{84.0} & \textbf{66.6} & \textbf{76.1} \\
+ & + & + & \textbf{82.6} & \textbf{66.1} & \textbf{77.2} \\
% \hline
% & & $E_\beta$ & 79.6 & 60.2 & 66.2 \\
\hline
% \multicolumn{3}{c||}{Evaluation on $E_\beta$} & 79.6 & 60.2 & 66.2 \\
% \hline
\end{tabular}
\end{center}
\caption{Ablation study: Relevance of different contributions. Each component is crucial for reaching the best performance. (Clust: $E_\beta$ training with clusters, Stand: standardization before clustering (Sec\ref{sec:surrogate}), MutInfo: mutual information loss (Sec\ref{sec:loss}))}
\label{tab:components}
\end{table}

\subsection{Testing Components and Parameters}\label{sec:ablations2}
In order to %study the effect of 
analyze our modules, we evaluate different models, each lacking one of the proposed contribution, see tab. \ref{tab:components}.
% after removing the components introduced by our approach (class standardization, mutual information minimization) and show the performances on standard benchmarks in Tab.\ref{tab:components}. 
% The table shows how each component from the proposed method is needed for the best performance. 
The table shows how each component is needed for the best performance. 
Comparing to the baseline in the first line, we see that simply introducing an additional task based on clustering the data deteriorates the performance, as we add another class-discriminative training signal that introduces worse or even contradictory information. However, by utilizing standardization, we allow our second encoder to explicitly learn new features to support the class encoder instead of working against it, giving a significant performance boost. A final mutual information loss emphasises the feature separation to improve the results further.

% Our approach can be combined with any existing metric learning loss. The effect of our model on other losses is evaluated in Tab.\ref{tab:cub} and \ref{tab:cars}.
Our approach can be combined with most existing metric learning losses, which we evaluate on ProxyNCA\cite{proxynca} and triplet loss with semihard sampling\cite{semihard} in Tab.\ref{tab:cub} and \ref{tab:cars}. On both CARS196 and CUB200-2011, we see improved image retrieval performance.
% We also show that our method is independent of architecture and choice of loss. 
% The different architectures and losses are compared in the corresponding tables. 
% Independent of the configuration, our approach successfully boosts the performances.

To examine the newly introduced hyper-parameters, Fig.\ref{fig:cluster_recall} compares the performances on the three benchmarks using a range of cluster numbers. The plot shows how the number of clusters influences the final performances, meaning the quality of the latent structure extracted by the auxiliary encoder $E_\beta$ is crucial for a better classification. At the same time, an optimal performance, within a range of $\pm1\%$ Recall@1, is reached by a large set of cluster values, making the model robust to this hyper-parameter. For these cumulative tests, a higher learning rate and less training epochs were used to both reduce computation time and avoid overfitting to the test set.
% the result shows that our model is robust with respect to this parameter, with some optimal range. 
Based on these examinations, we set a fixed, but dataset-dependent cluster number for all other training runs, see Sec. \ref{sec:impl}.

\begin{figure}[t]
\begin{center}
   \includegraphics[width=0.90\linewidth]{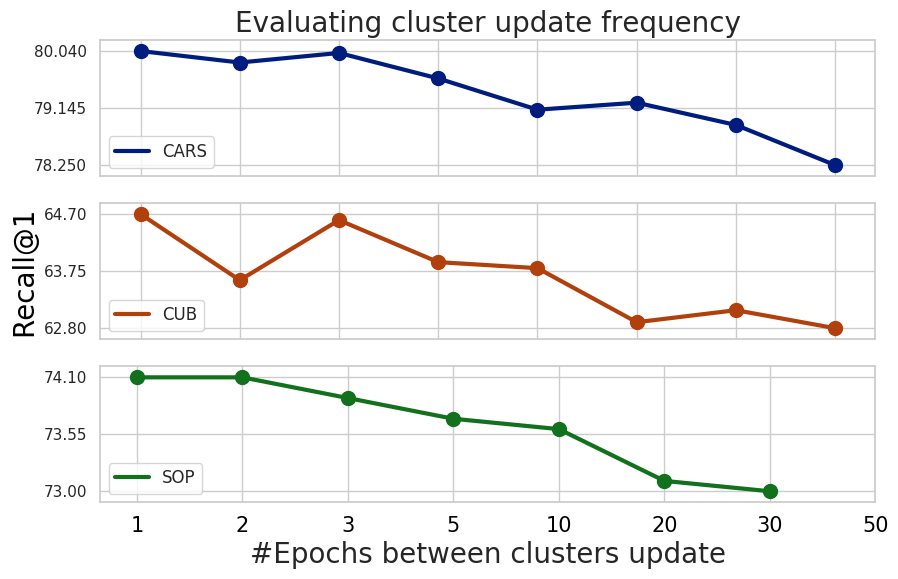}
\end{center}
   \caption{Ablation study: influence of the cluster label update frequency on Recall@1. An optimal number of clusters (see Sec. \ref{sec:impl}) and consistent scheduling was used.}
\label{fig:update_recall}
\end{figure}

A similar evaluation has been performed on the update frequency for the auxiliary labels (Fig.\ref{fig:update_recall}). Updating the cluster frequently clearly provides a boost to our model, suggesting that the auxiliary encoder $E_\beta$ improves upon the initial clustering. However, within a reasonable range of values (between an update every 1 to 10 epochs) the model has no significant drop in performance. Thus we fix this parameter to update every two epochs for all the experiments.

%-------------------------------------------------------------------------
\section{Conclusion}
In this paper we have introduced a novel extension for standard metric learning methods to incorporate structured intra-class information into the learning process. We do so by separating the encoding space into two distinct subspaces. One incorporates information about class-dependent characteristics, with the remaining encoder handling shared, class-independent properties. While the former is trained using standard metric learning setups, we propose a new learning task for the second encoder to learn shared characteristics and explain a combined training setup. 
Experiments on several standard image retrieval datasets show that our method consistently boost standard approaches, outperforming the current state-of-the-art methods and reducing intra-class variance.\\
\\
\textbf{Acknowledgements.} This work has been supported by Bayer and hardware donations by NVIDIA corporation.
%We thankfully acknowledge NVIDIA for their donation of Titan X GPUs used for this research. We also acknowledge Bayer for DOING SOMETHING SOMETHING.
%Experiments on several standard image retrieval datasets, namely CUB200-2011 \cite{cub200-2011}, CARS196 \cite{cars196}, Stanford Online Products \cite{lifted}, In-Shop Clothes \cite{inshop} and PKU VehicleID \cite{pku} show that our method consistently outperforms the current state-of-the-art methods.%, scaling with the available class variability.

\clearpage
{\small
\bibliographystyle{ieee_fullname}
\bibliography{main}
}

\end{document}